\documentclass[10pt,twocolumn,letterpaper]{article}

\usepackage{cvpr}
\usepackage{times}
\usepackage{epsfig}
\usepackage{graphicx}
\usepackage{amsmath}
\usepackage{amssymb}
\usepackage{algorithm}
\usepackage[noend]{algpseudocode}
\usepackage{pdfpages} 
\usepackage{booktabs}
\usepackage{enumitem}
\usepackage{layouts}
\usepackage{multirow}
\usepackage[dvipsnames]{xcolor}
\usepackage{scrextend}
\usepackage{bm}
\usepackage{colortbl}
\usepackage{sidecap}
\usepackage{footnote}
\usepackage{url}
\usepackage{titling}

\definecolor{color3}{rgb}{0.95,0.95,0.95}
\definecolor{color4}{rgb}{0.96,0.96,0.86}
\definecolor{color1}{rgb}{0.90,0.94,0.84}
\definecolor{color2}{rgb}{1,0.92,0.8}
\definecolor{unseenColor}{rgb}{0.91, 0.41, 0.17}

\usepackage[pagebackref=true,breaklinks=true,letterpaper=true,colorlinks,bookmarks=false]{hyperref}

\cvprfinalcopy % *** Uncomment this line for the final submission

\def\cvprPaperID{4510} % *** Enter the CVPR Paper ID here

\begin{document}

%%%%%%%%% TITLE
\title{\bf\Large\vspace{-1.2em}From Known to the Unknown: Transferring Knowledge to Answer \\ Questions about Novel Visual and Semantic Concepts}

\author{\large Moshiur R Farazi$^{1,2}$, Salman H Khan$^{2,3,1}$, Nick Barnes$^{1,2}$\\
\large $^{1}$Australian National University, $^{2}$Data61 - CSIRO, Australia, $^{3}$Inception Institute of AI, UAE\\
{\tt\small moshiur.farazi@anu.edu.au}
}

% \and
\date{\vspace{-0ex}} 

\maketitle
%\thispagestyle{empty}

%%%%%%%%% ABSTRACT
\begin{abstract}
Current Visual Question Answering (VQA) systems can answer intelligent questions about `Known' visual content. However, their performance drops significantly when questions about visually and linguistically `Unknown' concepts are presented during inference (`Open-world' scenario). A practical VQA system should be able to deal with novel concepts in real world settings. To address this problem, we propose an exemplar-based approach that transfers learning (i.e., knowledge) from previously `Known' concepts to answer questions about the `Unknown'. We learn a highly discriminative joint embedding space, where visual and semantic features are fused to give a unified representation. Once novel concepts are presented to the model, it looks for the closest match from an exemplar set in the joint embedding space.  This auxiliary information is used alongside the given Image-Question pair to refine visual attention in a hierarchical fashion. Since handling the high dimensional exemplars on large datasets can be a significant challenge, we introduce an efficient matching scheme that uses a compact feature description for search and retrieval. To evaluate our model, we propose a new split for VQA, separating Unknown visual and semantic concepts from the training set\footnote{Dataset and models will be available at: \url{TBA}}. Our approach shows significant improvements over state-of-the-art VQA models on the proposed Open-World VQA dataset and standard VQA datasets.
% Our approach shows significant improvements over state-of-the-art methods on VQAv1 and VQAv2 datasets using ours and previously used novel concept splits
%We demonstrate that the performance of current state-of-the-art VQA models drop significantly for unknown concepts. To address this problem, we propose a new transfer learning approach that learns to use the knowledge acquired on closely related exemplars to answer questions about unknown concepts. The knowledge acquired on Image-Question pairs is encoded using a rich joint embedding of visual and language features.  

\end{abstract}

%%%%%%%%% BODY TEXT
%=========================================================================
\section{Introduction}
Machine vision algorithms have significantly evolved various industries such as internet commerce, personal digital assistants and web-search. A major component of machine intelligence comprises of how well it can comprehend visual content. A Visual Turing Test to assess a machine's ability to understand visual content is performed with the Visual Question Answering (VQA) task. Here, machine vision algorithms are expected to answer intelligent questions about visual scenes.  The current VQA paradigm is not without its grave weaknesses. One key limitation is that the questions asked at inference time only relate to the concepts that have already been seen during the training stage (\emph{closed-world} assumption). 

%%%%%%%%%%%%%%%%%%%%%%%%%%%%%%%%
\begin{figure}[t]
\begin{center}
  \includegraphics[width=.9\linewidth]{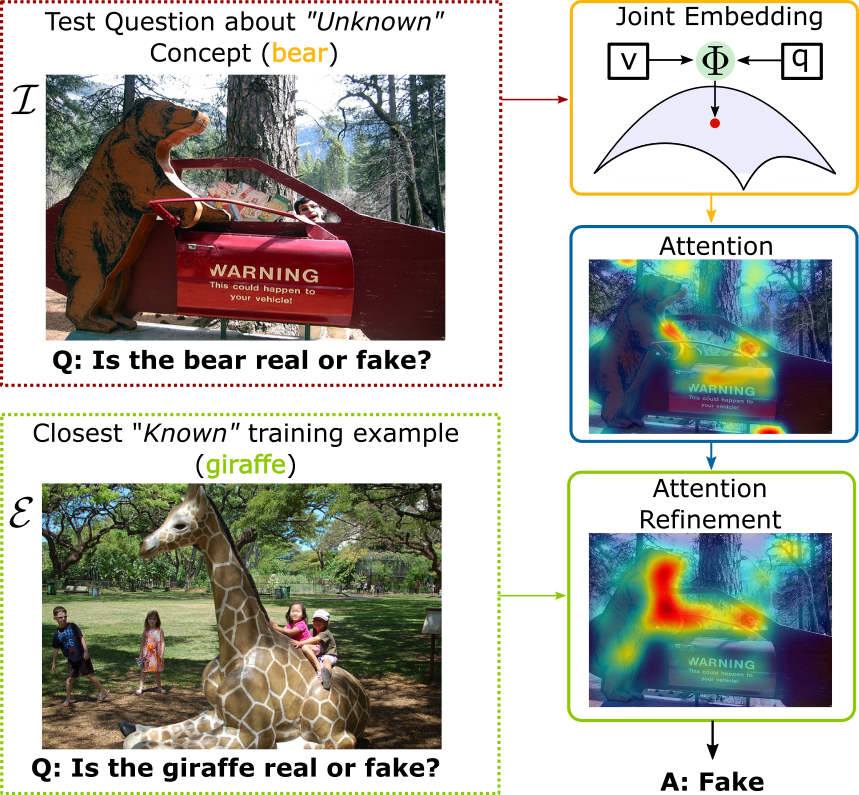}
\end{center}\vspace{-0.8em}
  \caption{Open World VQA for Novel Concepts: Our model learns to represent multi-modal information (Image ($\bm{I}$)-Question ($\bm{Q}$) pairs) as a joint embedding ($\Phi$). Once presented with novel concepts, the proposed model learns to effectively use past knowledge accumulated over the training set to answer intelligent questions. %$\mathcal{E}$ denotes an exemplar. 
  }
\label{fig:long}
\end{figure}
%%%%%%%%%%%%%%%%%%%%%%%%%%%%%%%

%\MRF{May be in the context of VQA, novel visual feature and sementic feature} 
In the real world, humans can easily reason about visually and linguistically \emph{Unknown} concepts based on previous knowledge about the \emph{Known}. For instance, having seen visual examples of '\emph{\textcolor{blue}{li}on}' and `\emph{ti\textcolor{blue}{ger}}', a person can recognize an unknown `\emph{\textcolor{blue}{liger}}' by associating visual similarities with a new compositional semantic concept and answer intelligent questions about their count, visual attributes, state and actions. In order to design machines to mimic human visual comprehension abilities, we must impart lifelong learning mechanisms that allow them to accumulate and use past knowledge to relate Unknown concepts. In this paper, we introduce a novel VQA problem setting that evaluates models in an `\emph{Open-World}' dynamic scenario where previously unknown concepts show up during the test phase (Fig.~\ref{fig:long}). 

An open-world VQA setting requires a vision system to acquire knowledge over time and later use it intelligently to answer complex questions about \emph{Unknown} concepts for which no linguistic+visual examples were available during training. Existing VQA systems lack this capability as they use a `fixed model' to acquire learning and envisage answers without explicitly considering closely related examples from the knowledge base. %\MRF{We are also not using knowledge base} \SK{The exemplar set is our knowledge base, I have clarified it in the next sentence} 
This can lead to `catastrophic forgetting' \cite{mccloskey1989catastrophic} as the object/question set is altered with updated categories. Here, we develop a flexible knowledge base (only comprising of the training examples) that stores the joint embeddings of visual and textual features. Our proposed approach learns to utilize past knowledge to answer questions about unknown concepts. 

Related to our work, we note a few recent efforts in the literature that aim to extend VQA beyond the already known concepts \cite{wang2017vqa, Agrawal_2018_CVPR, teney2016zero, ramakrishnan2017empirical, agrawal2017c}. A major limitation of these approaches is that they introduce novel concepts only on the language side (\ie, new questions/answers), either to re-balance the split or to prevent the model cheating by removing biases \cite{Agrawal_2018_CVPR, teney2016zero, agrawal2017c}. Further, they rely on external data sources (both visual and semantic) and consider training splits that contain visual instances of `novel objects', thereby violating the unknown assumption \cite{ramakrishnan2017empirical, teney2016zero}. 
To bridge this gap, we propose a new Open World VQA (OW-VQA) protocol for novel concepts based on MS-COCO and VQAv1-v2 datasets. Our major contributions are:

\begin{itemize}[topsep=0pt,noitemsep]
    \item We reformulate VQA in a transfer learning setup that uses closely related \emph{Known} instances from the exemplar set to reason about \emph{Unknown} concepts. 
    \item We present a novel network architecture and training schedule that maintains a knowledge base of exemplars in a rich joint embedding space that aggregates visual and semantic information. 
    \item We propose a hierarchical search and retrieval scheme to enable efficient exemplar matching on a high dimensional joint embedding space.
    \item We propose a new OW-VQA split to enable impartial evaluation of VQA algorithms in a real-world scenario and report impressive improvements over recent approaches with our proposed model.
\end{itemize}

%%%%%%%%%%%%%%%%%%%%%%%%%%%%%%%%
\begin{figure*}
\begin{center}
% \fbox{\rule{0pt}{2in} \rule{.9\linewidth}{0pt}}
\includegraphics[width=\linewidth]{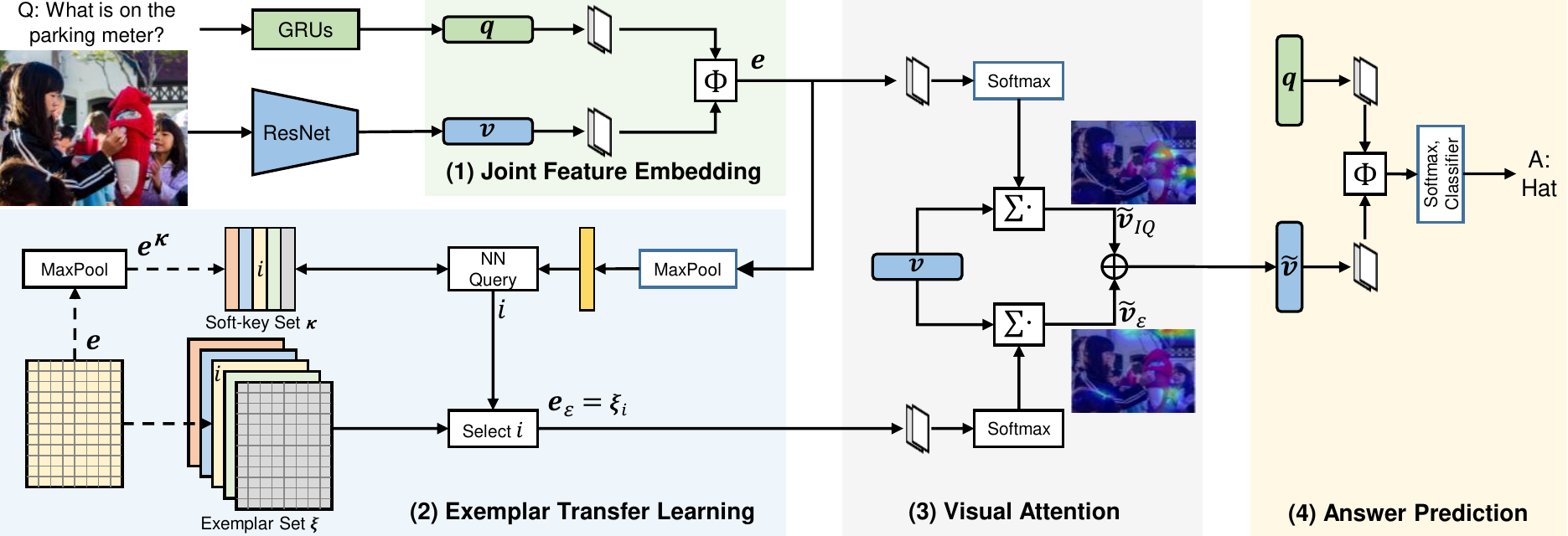}
\end{center}\vspace{-0.7em}
  \caption{Overview of our proposed Joint Embedding Exemplar (JEX) model. Given an Image-Question (IQ) pair, our model populates an exemplar set in joint embedding space (represented by dotted lines) during training. When novel concepts (both visual and semantic) are encountered in testime, our model identifies closely related exemplar in the joint embedding space to attend to most relevant image details which in turn is used to generate an intelligent answer.}%\MRF{Figure generated from Fig 2 pipeline.pptx in the image folder}
  %\NB{You've dropped $\tau$? }\SK{$\tau$ is the parameter tensor for $\Phi$. It is represented within $\Phi$. Similarly $\tau_v$ and others are matrices that are not inputs (previously they were shown as inputs to $\Phi$).}
\label{fig:pipleline}
\end{figure*}
%%%%%%%%%%%%%%%%%%%%%%%%%%%%%%%%

%=========================================================================
\section{Related Works}

\textbf{Multi-modal Information Fusion:} VQA is an AI complete task that requires high-level multimodal reasoning both in visual and language domains.  
%\NB{remained: not the right word - not sure what you mean}
%In the recent literature, a strong focus has on the optimal mechanism to fuse multimodal cues. 
The recent literature in VQA mostly focus on the optimal mechanisms to fuse multimodal cues. A simple fusion approach was used by Lu \etal \cite{lu2016hierarchical} that progressively combines multimodal features using concatenations and sum-pooling operations. Xu \etal \cite{xu2015show} proposed a recurrent neural network to generate intelligent image captions by considering the previously predicted words and targeted visual content. Bilinear models provide an effective way to model complex interactions, but impose restrictions due to computational intractability for high-dimensional inputs. Efficient versions of bilinear pooling were used in \cite{gao2016compact, kim2016hadamard} to learn the second-order interactions between visual and language features. To further speed-up the computations, Ben-younes \etal \cite{benyounescadene2017mutan} introduced a Tucker fusion scheme that first projects the individual modalities to low dimensions and subsequently learns full bilinear relationships. Recently, Farazi \etal \cite{Farazi_2018_BMVC} fused complementary object level features alongside image level descriptors to achieve superior performance. 

\textbf{VQA for Novel Concepts:} 
%   VQA  engine  is  highly
% likely to encounter questions about totally unseen objects.
% Some  recent  attempts  to  propose  novel  concept  splits  for
% VQA  that  only  consider  the  language  side  [1,  23,  20,  2].
% E.g.,  [1]  showed  that  existing  VQA  datasets  have  highly
% correlated answers on train and test sets.  As a result, VQA
% models tend to remember the popular answers instead of at-
% tending to correct image details for correct answer.  They
% subsequently proposed new protocols with distinct distribu-
% tions of answers in both sets to have a fair evaluation pro-
% tocol. On similar lines, [2] proposed a new split to test new
% compositions  of  seen  concepts  in  questions  and  answers
% (Q&A). Teney and Hengel [23] also highlighted that cur-
% rent VQA models are biased towards rare and unseen con-
% cepts and proposed a zero-shot split only for language con-
% tent (
% i.e
% ., Q&A). We note that the above-mentioned methods
% only suggest a language based split and the visual concepts
% may still appear visually during the training process. There-
% fore, they do not satisfy the zero-shot assumption
%The zero-shot learning (ZSL) paradigm allows a model to reason about objects that we have not seen before \cite{xian2018zero, rahman2018zeroshot}. 
A VQA engine is highly likely to encounter questions about totally unknown objects. Some recent attempts to propose novel concept splits for VQA only consider the language side \cite{Agrawal_2018_CVPR, teney2016zero, ramakrishnan2017empirical, agrawal2017c}. Goyal \etal \cite{goyal2016making} showed that existing VQA datasets have highly correlated answers on train and test sets. As a result, VQA models tend to remember the popular answers instead of attending to correct image details for correct answer. They subsequently proposed new protocols with distinct distributions of answers in both sets to have a fair evaluation protocol. On similar lines, Agrawal  \etal \cite{Agrawal_2018_CVPR} proposed a new split for VQA where train and test sets have different prior distributions for each question type. Teney and Hengel \cite{teney2016zero} also highlighted that current VQA models are biased towards rare and unseen concepts and proposed a zero-shot split only for language content (\ie, Q\&A). We note that the above-mentioned methods only suggest a language based split and the visual concepts may still appear visually during the training process. Therefore, they do not satisfy the zero-shot assumption.  
%To apply a VQA model to new datasets, a domain adapted version for the VQA task was recently proposed in \cite{chao2018cross}. However, they assumed that the distribution for novel datasets is known apriori and do not consider a zero-shot split. 

\textbf{Exemplars for VQA:} 
Although most VQA approaches only work with the given training set, some efforts explore the use of supplementary information to help the VQA system. Generally, such methods employ external knowledge sources (both textual and visual) to augment the training set. For example, Teney \etal in \cite{teney2016zero,teney2017tips} used web searches to find related images which were used for answer prediction. Language based external knowledge bases were used by Wang \etal\cite{wang2017explicit} and Wu \etal\cite{wu2016ask} to provide logical reasons for each answer choice and to answer a more diverse set of questions. More recently, Teney \etal\cite{Teney_2018_ECCV} proposed a meta-learning approach that learns to use an externally supplied support set comprising of example questions-answers. In contrast to these approaches, we do not use any external data, rather learn an attention function that learns to use similar examples from the training set to provide better inference-time predictions. Patro \etal \cite{patro2018differential} proposed a differential attention mechanism that uses an exemplar from the training set to generates human-like attention maps, however does not consider a transferable attention function that can reason about new visual/semantic concepts.

%=========================================================================
\section{Methods}
Given a question $\bm{Q}$ about an image $\bm{I}$, an AI agent designed for the VQA task will predict answer $a^*$ based on the learning acquired from training examples. This task can be formulated as: 
\begin{align}
    a^* = \underset{\hat{a} \in \mathcal{D}}{\arg\max}\, P(\hat{a}|\bm{Q},\bm{I}; \bm{\theta}) 
\end{align}
where $\bm{\theta}$ denotes the model parameters and $a^*$ is predicted from the dictionary of possible answers $\mathcal{D}$. An ideal VQA system should effectively model the complex interactions between the language and visual domains to acquire useful knowledge and use it to answer newly presented questions at test time. Towards this end, we propose a  framework to answer questions about novel concepts in Fig.~\ref{fig:pipleline}. The overall pipeline is based on four main steps: \emph{(1) Joint Feature Embedding:} The given IQ pair is processed to extract visual $\bm{v}$ and language $\bm{q}$ features. These features are jointly embedded into a common space through multi-modal fusion.
%techniques ranging from concatenation, element wise product or sum to multimodal pooling, trucker fusion. 
\emph{(2) Transfer Learning with Exemplars:}  We propose an exemplar-based model that learns to reason from similar examples known during training time. When presented with a test image containing an Unknown concept, our model transfers the knowledge acquired on closely related examples of Known concepts to novel cases. \emph{(3) Visual Attention:} In the next stage, the proposed network selectively attends to visual scene details using the joint feature embedding of given inputs and the exemplars (output of step $1$ and $2$ respectively). This ensures that the model learns to identify salient features to answer a specific question. \emph{(4) Answer Prediction:} During the final stage, the refined joint embedding is calculated by the model to predict the correct answer $a^{*}$ from the answer set $A$, minimizing a cross entropy loss. 
% Next, we describe the individual components of our method.

%-------------------------------------------------------------------------
\subsection{Joint Feature Embedding}
\label{sec:joint_em}
%\NB{Need to define $n_v,n_q,n_e$}
From the given image $\bm{I}$, the $n_v$-dimensional visual feature embedding $\bm{v}\in\mathbb{R}^{n_v}$ is extracted from the last convolutional layers just before the global pooling and classification layer of the feature extraction model (\ie ResNet \cite{he2016deep}). The language feature embedding $\bm{q}\in\mathbb{R}^{n_q}$ is generated from $\bm{Q}$ by first encoding the question in a one-hot-vector representation and then embedded into vector space using Gated Recurrent Units (GRUs) 
%\NB{write out fully the first time you use it} 
\cite{cho2014properties, fukui2016multimodal}. 
In order to predict a correct answer, a VQA model needs to generate a joint embedding: $\bm{e} = \Phi(\bm{v}, \bm{q}; \bm{\tau}) \in \mathbb{R}^{n_e}$. A naive approach that models visual-semantic interactions using a tensor $\bm{\tau} \in  \mathbb{R}^{n_q \times n_v \times n_e}$ will result in an unrealistic number of trainable parameters (\eg, $\sim9.83$ billion for our baseline model). 

To reduce the dimensionality of the tensor, we use Tucker decomposition \cite{tucker1966some} which can be seen a high-order principal component analysis operation. This technique has been proven effective in embedding visual and textual features for VQA \cite{Farazi_2018_BMVC, benyounescadene2017mutan}. It approximates $\bm{\tau}$ as follows:
\begin{align}\label{eq:tucker}
    \bm{{\tau}} &= \sum_{i=1}^{t_q} \sum_{j=1}^{t_v} \sum_{k=1}^{t_e} \omega^{ijk} \tau^i_q \circ \tau^j_v \circ \tau^k_e \notag\\
    &= \bm{\omega} \times_{1} {\bm{\tau}}_q \times_{2} {\bm{\tau}}_v \times_{3} {\bm{\tau}}_e = [\![\bm{\omega}; {\bm{\tau}}_q, {\bm{\tau}}_v , {\bm{\tau}}_e]\!]
\end{align}
%\NB{What is the meaning of notation $\times_1$ etc}
where $\times_i$ denotes n-mode product of a tensor with a matrix and $\circ$ denotes the outer vector product. Eq.~\ref{eq:tucker} means that tensor $\bm{\tau}$ is decomposed into a core tensor $\bm{\omega} \in \mathbb{R}^{t_q \times t_v \times t_e}$ and orthonormal factor matrices ${\bm{\tau}}_q \in \mathbb{R}^{n_q \times t_q}$, ${\bm{\tau}}_v \in \mathbb{R}^{n_v \times t_v}$, ${\bm{\tau}}_e \in \mathbb{R}^{n_e \times t_e}$. Intuitively, by setting $t_q < n_q$, $t_v < n_v$ and $t_e < n_e$ of the factored matrices, one can approximate $\bm{\tau}$ with only a fraction of originally required number of trainable parameters.

The output embedding $\bm{e}$ from the Tucker decomposition effectively captures the interactions between semantic and visual features for a given Image-Question-Answer (IQA) triplet. Such joint embedding for VQA is specific to the given IQ pair because the same visual feature associated with different semantics (and vice-versa) will result in a different joint embedding specific for that pair. For example, given an image that captures children playing in the backyard, when asked \textit{`How many children are in the picture?'} and \textit{`Are the children happy?'}, requires two very different joint embeddings $\bm{e}$ even though they use the same visual feature, $\bm{v}$. Building on this rich joint embedding, we develop a transfer learning module based on exemplars. 

%-------------------------------------------------------------------------
%-------------------------------------------------------------------------
\subsection{Exemplar based Learning Transfer}
\label{sec:ex_em}
 Given a question about an Unknown concept, our model identifies a similar joint embedding of \emph{Known} concepts from the training set. Since,  visual/semantic examples of unknown concepts are not available to us during training, first we learn a generic attention function $\mathcal{A}$ that can transfer knowledge from the Known concepts to Unknown. The attention function is learned on the training set, where it identifies the useful features from closely related exemplars to answer questions. The function $\mathcal{A}$ is agnostic to specific IQ pairs and provides a generalizable mechanism to identify relevant information from related examples. Therefore, at inference time, we use the same exemplar based attention function to obtain refined attention maps by using the closely related joint embedding of known concepts. We design the training schedule in two stages to facilitate knowledge transfer. During the first stage, only the Visual-Semantic embedding part of the network is trained end-to-end and the joint feature embedding tensor $\bm{e}$ is stored in memory $\bm{\xi} \in \mathbb{R}^{d \times n}$, where $n$ is the number of training IQA triplets and $d$ denotes the embedding dimension. %\NB{Should this all be in 3.1?} 
In the second stage, both the visual-semantic embedding and the exemplar-embedding segment of the model are trained end-to-end where the model performs a nearest neighbour (NN) search on $\bm{\xi}$ to find the most similar joint embedding $\bm{e}_{\xi}$. Further, the network learns to use the exemplar embedding to refine the attention on visual scene details. 
  %and \NB{learns to make make use of the additional knowledge : what does  this mean?}.  
This can be represented as:
  \begin{align}
      \bm{\tilde{v}}_{\mathcal{E}} = \mathcal{A}(\bm{v},\bm{e}_{\mathcal{E}}),\text{ where, } \bm{e}_{\mathcal{E}} = \mathcal{N}(\bm{e}, \bm{\xi}, \bm{\kappa}),
  \end{align}
where, $\bm{e}_{\mathcal{E}}$ is the exemplar-embedding found using nearest neighbour search ($\mathcal{N}$) on a set of compact embeddings  $\bm{\kappa}$.

There are two main motivations for not performing the NN search directly on $\bm{\xi}$ and instead using a set of compact embeddings  $\bm{\kappa}$. Firstly, searching in the joint embedding space would allow the model to overfit when searching for the closest match. However, when searching for the closet match for a compact representation of the joint embedding, the reduced dimensionality of the representation avoids overfitting. Secondly, storing and performing NN search directly on the joint embedding exemplar space is extremely memory and time intensive. For example, if visual features are extracted from the second-last convolution layer of ResNet152\cite{he2016deep} for evaluation on VQAv2\cite{goyal2016making} dataset, $\bm{\xi}$ will be $\mathbb{R}^{n \times d}$ dimensional where $n\approx 400K$ training examples and $d = t_e \times G$ such that grid locations $G=14\times14$ and $t_e \approx 500$ for a standard setting. Doing a similarity match on such a large space has practical memory and computational limitations.

%\NB{Any particular reason why max pool for this reduction?} 
Due to the above-mentioned reasons, we generate a coarse representation of $\bm{\xi}$ by passing each of its elements through a max-pooling layer. We empirically found max-pooling to perform well in our case. The set of max-pooled embeddings is represented by $\bm{\kappa}$, whose entries act as \emph{soft-keys} for the exemplar-embeddings. When a query embedding $\bm{e}$ is presented, we calculate its compact feature $\bm{e}^{\kappa}$ by applying a max-pooling operation. The NN search is performed between $\bm{e}^{\kappa}$ and each element of $\bm{\kappa}$ to find the embedding $\bm{e}^{\kappa}_{\mathcal{E}}$. As the elements of $\bm{\xi}$ and $\bm{\kappa}$ have a one-to-one relationship, by matching the maxpooled version of $\bm{e}$ to $\bm{\kappa}$, the model finds the exemplar embedding $\bm_{\mathcal{E}}$. Notably, with this setup, we do not require the large set of exemplars $\bm{\xi}$ to be loaded into memory, instead a much more compact representation is used for efficient search and retrieval.

%-------------------------------------------------------------------------
\subsection{Visual Attention}
\label{sec:co_att}
Attention is applied at two levels in our network. At the first level, the joint embedding $\bm{e}$ is used to apply attention on visual features to focus model attention on more important features. The joint embedding is passed through a {FC} (fully connected) layer followed by a {softmax} layer to generate an attention vector $\bm{\alpha}_{IQ}^{v} \in \mathbb{R}^{G}$ which scores each spatial grid location $G$ of visual feature $\bm{v}$ according to input IQ pair. At the second level,  we select $\bm{e}_{\mathcal{E}}$, the most similar exemplar embedding of $\bm{e}$, and follow the same protocol to generate another attention score $\bm{\alpha}^v_{\mathcal{E}}$ over spatial grid locations. 
The attention vectors $\bm{\alpha}_{IQ}^{v}$ and $\bm{\alpha}^v_{\mathcal{E}}$ signify complementary proposals generated using a given IQ pair and the most similar visual-semantic embedding from the exemplar set respectively. Such a complementary attention mechanism allows the model to reason about unknown concepts using the attention calculated from the combined effect of the input IQ pair and further refine it by looking at the closest example from the exemplar set. Both the IQ pair and the exemplar-based attention vectors are used to take a weighted sum at each location $g$ of the input visual representation $\bm{v}$ (\ie, $\bm{v}^{g}$) to create an attended visual representation. This can be formulated as:
\begin{equation}
    \bm{\Tilde{v}}_{IQ} = \sum_{g=1}^{G} \bm{v}^{g} \bm{\alpha}^{v,g}_{IQ}  \quad \textrm{and} \quad
    \bm{\Tilde{v}}_{\mathcal{E}} = \sum_{g=1}^{G} \bm{v}^{g} \bm{\alpha}^{v,g}_{\mathcal{E}},
\end{equation}
where $\bm{\Tilde{v}}_{IQ} $ and $\bm{\Tilde{v}}_{\mathcal{E}} $ denote the attended visual features generated from the IQ pair and exemplar embedding respectively.
%The attended visual features represents the salient parts of the image that were deemed important by the input image-question pair and most similar training example. Both of these encapsulate task specific visual information required to predict the correct answer.
We concatenate the two to create an overall attended visual feature $\bm{\Tilde{v}}$ and again apply $\Phi$ in a similar manner as described in Sec.~\ref{sec:joint_em} to generate the final vision-semantic embedding. We then project the embedding to the prediction space that is passed through a classifier to predict the final answer $a^{*} \in \mathcal{D}$. %\NB{Should this been expressed in Fig 2?}\SK{The above pipeline is shown in Fig. 2, you mean we include 'D' as well?}
%We then project the embedding to the prediction space and generate $a \mathbb{R}^{n_a}$ first by passing the embedding through a fully connected layer and then subjecting it to $softmax$ transformation. $a$ is then passes through the classifier that predicts $\Tilde{a}$ for a given question $q$ and image $v$.

%-------------------------------------------------------------------------

%=========================================================================
\section{OW-VQA dataset generation protocol}
\label{sec:dataset_gen}
\textbf{Motivation:} 
%A VQA system learns to predict an answer based on the example IQ pairs that it has seen during the training. 
When a VQA system is subjected to an open-world setting, it can encounter numerous visual and semantic concepts that it has not seen during training. 
%\NB{These two sentences are a long introduction here given its a major message of the paper} 
To help VQA systems develop capability to handle unknown visual and semantic concepts, we propose a new split that contains known/unknown concepts for training/testing respectively. Our dataset generation protocol builds on the fact that images in VQA datasets \cite{antol2015vqa,goyal2016making} are repurposed from MSCOCO images \cite{lin2014microsoft} and paired with crowd sourced Q\&A. Even though, MSCOCO images have rich object level annotation for $12$ super-categories and $80$ object categories, the VQA dataset annotations include only information related to Q\&A, excluding any link to object level annotation. This constitutes a significant knowledge gap which if addressed, would allow for more subtle understanding of the scene even if it contains previously unknown visual and semantic concepts. To bridge the gap, we propose to use Objects categories as the core entity to develop a true Known-Unknown split that constitutes both visual and semantic domains. First, we propose an known-unknown  split for the MSCOCO object categories, which leads to a well-founded split that separates known/unknown concepts in IQA triplets on VQA datasets.

\begin{figure}[!tp]
\begin{center}
% \includegraphics[trim = 0mm 0mm 0mm 6mm, clip, width= 0.9\linewidth]{images/Fig_stacked_bar_ver.pdf}
\includegraphics[width= .8\linewidth]{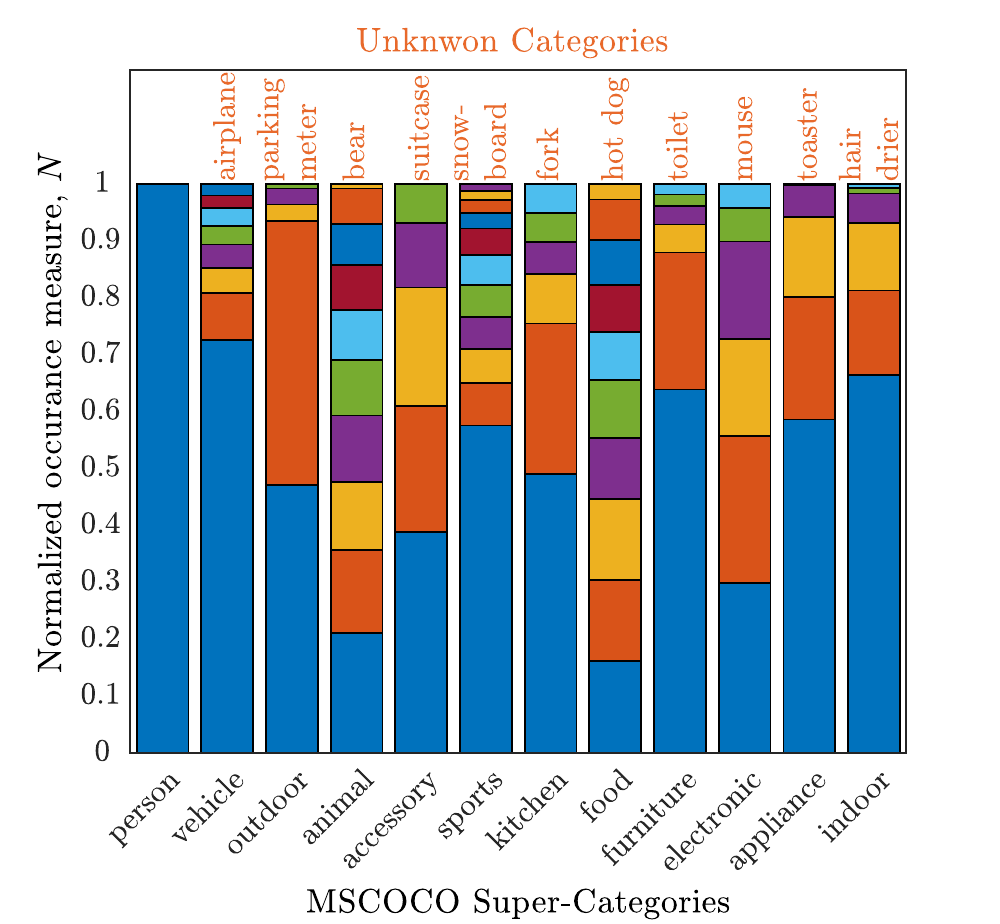}
\end{center}
\vspace{-1.0em}
  \caption{Selected \emph{\textcolor{unseenColor}{Unknown}} categories from each super-category of MSCOCO dataset based on least occurrence measure, $N$. %\MRF{Figure generated from fig stacked bar unseen.m in images folder}
  }
\label{fig:stacke_bar}
\end{figure}

\textbf{Known/Unknown Object Split:} 
At the first stage, from each MSCOCO super-category (except for {\tt\small person} which has no sub-category), we select the rarest category as \textit{Unknown} and the rest as \textit{Known}. This choice is motivated by the fact that rare classes are most likely to be unknown. For each category $c$, we calculate $N_{i}$ and $N_{t}$ which represent the total number of images that $c$ appears in, and total number of instances of $c$ respectively. We define $N = N_{i} \times N_{t}$ as the measure of occurrence for each category. We select the category with the smallest $N$ as Unknown category. Fig. \ref{fig:stacke_bar} shows the normalized $N$ for categories in each super category and respective  \emph{\textcolor{unseenColor}{Unknown}} categories (more details in supplementary material). This ensures that the unknown category appears in least number of images least number of times. Such a measure is particularly necessary for datasets that are used to perform high level vision tasks associated with a language component. For example in super-category {\tt\small vehicle}, {\tt\small train} is less frequent compared to {\tt\small airplane} in terms of instances ($4{,}761$ vs $5{,}278$). If the split was solely based on number of instances $N_{t}$, then {\tt\small train} would have been selected as an unseen class even though it appears in $662$ less number of images than {\tt\small airplane}. When human annotators are tasked with generating language components (\ie Q\&A or captions), the rarest language cues are often associated with categories that appears in the least number of images, the least number of times. Thus, selecting the category with the least occurrence measure $N$ ensures that categories with least language representation are selected as \textit{Unknown}. 
% Fig.~\ref{fig:stacke_bar} shows the selected unseen class from the dataset. 
% we calculate $N_{i}$ (the total number of images where $c$ appears at least once) and $N_{t}$ (the total number of instances of $c$ in both \textit{train2014} and \textit{val2014} splits of MSCOCO). 
% We define $N = N_{i} \times N_{t}$  and 
 
%\NB{This example is not clear what you are saying}
%This is due to the fact when picture of an airplane in taken, most time more that one instance of airplane in evident and such natural imbalance is expected across all object categories. 
%%%%%%%%%%%%%%%%%%%%%%%%%%%%%%%%
%%%%%%%%%%%%%%%%%%%%%%%%%%%%%%%%%%%%%%%
%%%%%%%%%%%%%%%%%%%%%%%%%%%%%%%%%%%%%%%
\begin{table}[!tp]
  \centering
  \scalebox{0.95}{
  \begin{tabular}{cc r r  r}
  \toprule \rowcolor{color3}
  \multicolumn{2}{l}{Dataset ($\rightarrow$)}         & COCO    & \multicolumn{1}{c}{OWv1}  
  & \multicolumn{1}{c}{OWv2}  \\
  \rowcolor{color3}
  \multicolumn{2}{l}{Split ($\downarrow$)}                 & \multicolumn{1}{c}{\# Image}  
  & \multicolumn{1}{c}{\# IQA}    & \multicolumn{1}{c}{\# IQA}    \\
  \midrule
  \multicolumn{2}{l}{Trainset}             &  69,557   & 187,986   & 336,124   \\
  \midrule
  \multirow{2}{*}{Valset}       & Known     &  34,117   & 120,916   & 178,321   \\ \cmidrule{2-5}
                                & Unknown   &  6,367    & 19,101    & 34,945    \\
  \midrule
  \multicolumn{2}{l}{Testset}              &  13,226   & 36,054    & 66,568    \\
  \bottomrule
  \end{tabular}}
  \vspace{1em}
  \caption{Split-wise data statistics for the three datasets used in our propose OW-VQA splits.}
  \label{tab:OW_dataset_split}
\end{table}
%%%%%%%%%%%%%%%%%%%%%%%%%%%%%%%%%%%%%%%

%%%%%%%%%%%%%%%%%%%%%%%%%%%%%%%%%%%%%%%
\begin{table*}[]
 \begin{minipage}[b]{0.56\textwidth}
  \centering
  \scalebox{0.8}{
  \begin{tabular}{l|cccc|cccc}
    \toprule
    		\rowcolor{color3}							& \multicolumn{4}{c|}{OW-VQAv1}  & \multicolumn{4}{c}{OW-VQAv2}\\
									
    			\rowcolor{color3}								& All 	& Y/N   & Num.  & Other & All 	& Y/N   & Num.  & Other \\ 
    \midrule
    JEX (Ours)                          & \textbf{61.7} & \textbf{81.2} & \textbf{40.3} & \textbf{48.5} 
                                        & \textbf{57.8} & \textbf{74.8} & \textbf{37.1} & \textbf{47.6}\\
    \midrule
    MUTAN \cite{benyounescadene2017mutan}
                                        & 60.3 & 80.6 & 39.5 & 48.1 & 56.9 & 74.7 & 36.2 & 46.9  \\
    MCB \cite{fukui2016multimodal}      & 59.7 & 73.1 & 36.9 & 46.1 & 55.5 & 71.8 & 35.5 & 45.7 \\
    SAN \cite{yang2016stacked}          & 55.7 & 76.0 & 40.2 & 39.8 & 50.6 & 67.2 & 34.5 & 39.4 \\
    HieCoAtt \cite{lu2016hierarchical}  & 55.6 & 77.3 & 42.1 & 37.7 & 50.7.& 67.4 & 35.1 & 38.5  \\
    % d-LSTM Q+norm I\cite{antol2015vqa}  & 54.1 & 77.3 & 37.2 & 35.9 & 49.8 & 68.1 & 37.1 & 35.7 \\
    VQA \cite{antol2015vqa}             & 54.1 & 77.3 & 37.2 & 35.9 & 49.8 & 68.1 & 37.1 & 35.7 \\
    \bottomrule
  \end{tabular}}\vspace{0.5em}
  \caption{Evaluation on our proposed OW-VQA split when trained on Trainvalset (Trainset+Valset-Known) and evaluated on Testset.
  }
  \label{tab:Test_on_zstest}
  \end{minipage}
  \hfill
  \begin{minipage}[b]{0.4\textwidth}
  \centering
  \centering
\includegraphics[width= .9\linewidth]{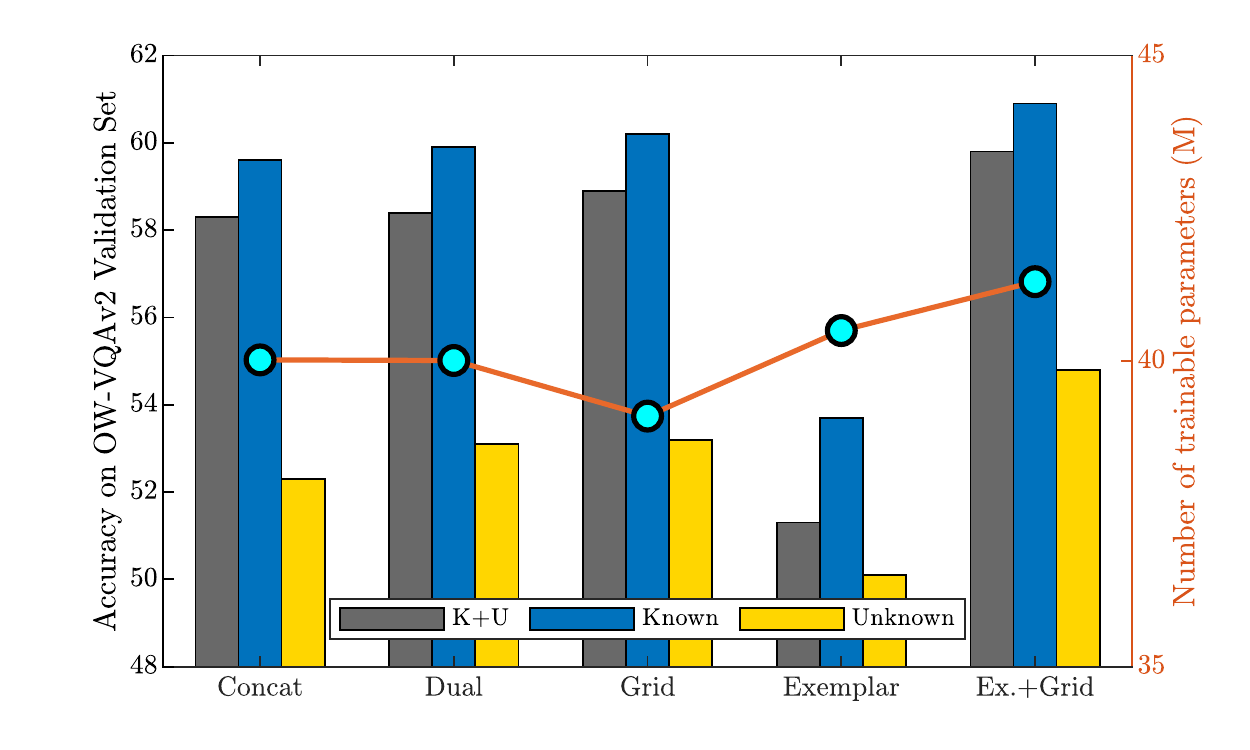}
\vspace{-1em}
  \caption{Performance drop when trained on known concepts and validated on unseen concepts.}
%   ?\SK{make it horiz.} \MRF{Tried to. Couldn't find barh to have 2 xaxis ticklable. May need to write custom script. Redid the figure with more width. }}
\label{fig:perfomance_drop_ablation}
  \end{minipage}\vspace{-0.5em}
\end{table*}
%%%%%%%%%%%%%%%%%%%%%%%%%%%%%%%%%%%%%%%

\textbf{Image-Question-Answer Split:} Building on the Known-Unknown categories, we repurpose IQA triplets from VQAv1\cite{antol2015vqa} and VQAv2\cite{goyal2016making}, and propose training (known) and test (unknown) splits called OW-VQAv1 and OW-VQAv2. For this purpose, we combine training and validation sets of respective datasets (test split cannot be used as they are not publicly available). We use two steps to ensure that both visual and semantic concepts associated with the Unknown categories are completely absent in the training set.
%in two steps. 
Firstly, we place an IQA triplet in the training set if there is no instance of any unseen category in the image of the corresponding triplet. This ensures that the new visual concepts are unknown to the model during training. Secondly, we focus on the semantic part and filter out the IQA triplets from the training set that have any unknown category names or synonyms in the questions. 
%This ensures that even though an unseen category is not present visually, the training set also does not contain any semantic cues of the unseen classes that model might learn during training. 
Such visual and semantic confinement of concepts in train/test split is the major advantage that our proposed dataset has over other approaches \cite{ramakrishnan2017empirical, teney2016zero, Agrawal_2018_CVPR} where the unseen `objects/concepts' are only defined at semantic-level. For example, {\tt\small{airplane}} is an `unseen category' in our proposed dataset and a `novel object' in the dataset proposed by Ramakrishnan \etal \cite{ramakrishnan2017empirical}. A semantically motivated dataset generation protocol would place an IQA triplet that does not have the keyword {\tt\small{airplane}} in the question, in the training set. However, there are several IQA triplets in VQA dataset that shows an {\tt\small{airplane}} being serviced by a car, truck or a person at an airport, and do not ask about the {\tt\small{airplane}}.
%even if the associated question does not mention `{\tt\small{airplane}}'.
%\SK{Can we show here one associated example here to illustrate this above example?}\MRF{Add to the qualitative results figure? Or a separate figure?} \SK{A small separate fig showing tis problem for [22] split.}. 
Just ensuring that the semantic concepts are not present during training only addresses a naive version of the challenge an open-world VQA system would face.

% This is particularly important when testing a model in for open-world setting, as its equally important to know how the model performs when tasked with known and unknown concepts.

%=========================================================================
\section{Experiments}
In this section, we first describe the experimental setup and implementation details of our proposed model. Then we present the baseline model architectures useing different combinations of visual and semantic features to generate joint embedding. Then we present the results of our experiments which includes benchmarking of VQA models on OW-VQA dataset, ablation and performance analysis of our proposed model on semantically motivated VQA splits and standard VQA setting.

%%%%%%%%%%%%%%%%%%%%%%%%%%%%%%%%
\begin{figure*}
\begin{center}
% \fbox{\rule{0pt}{4.5in} \rule{.95\linewidth}{0pt}}
\includegraphics[width=0.85\linewidth]{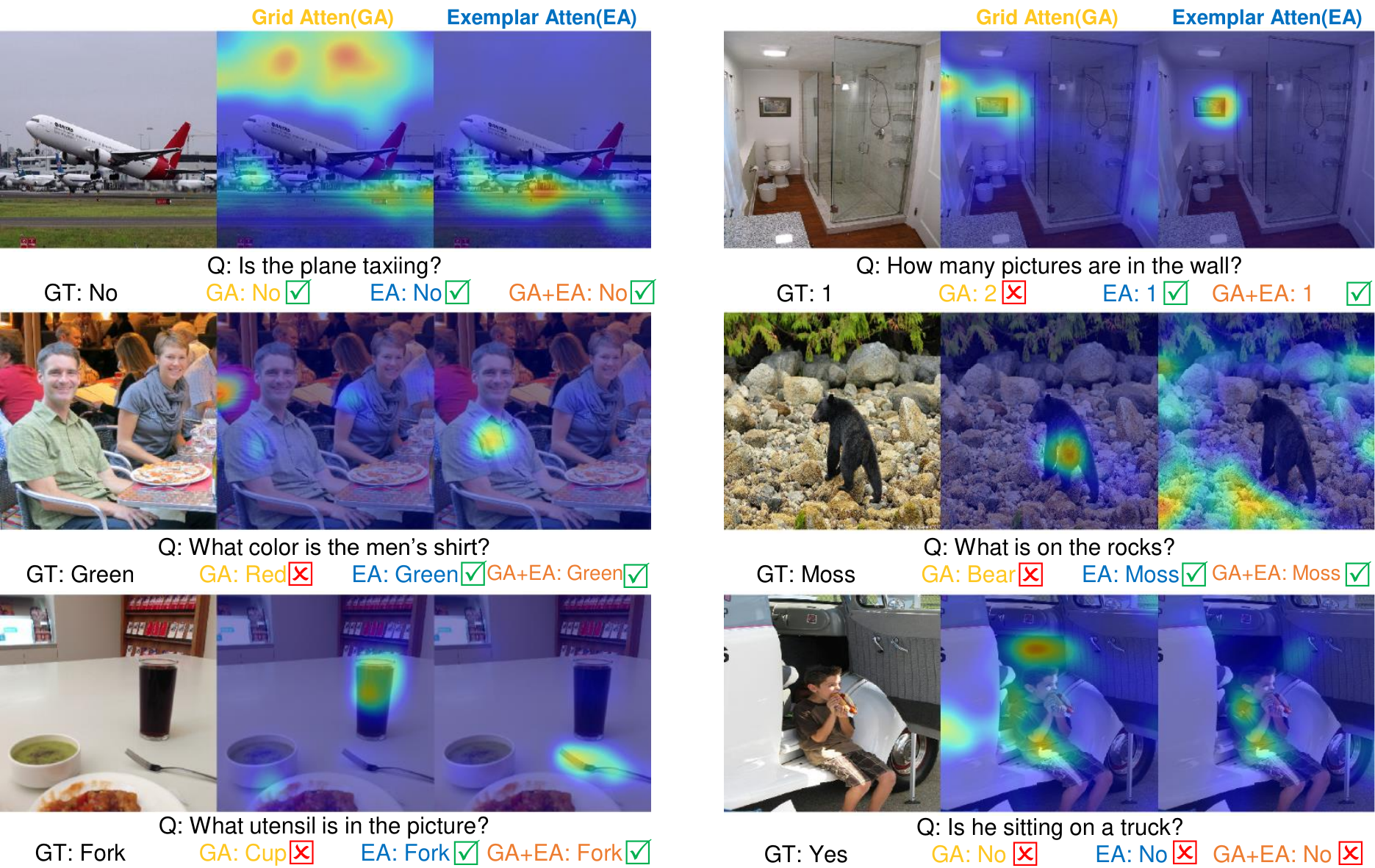}
\end{center}\vspace{-1em}
  \caption{Predicted answers and attention maps evaluting JEX model on OW-VQAv2 Valset-Unknown images. The Grid Attention map (GA) and Exemplar Attention(EA) map provides complementary information for the model to reason about Unknown concepts, where only using GA or EA often leads to wrong prediction.}
\label{fig:qual_results}\vspace{-0.5em}
\end{figure*}
%%%%%%%%%%%%%%%%%%%%%%%%%%%%%%%%
%-------------------------------------------------------------------------
\subsection{Experimental Setup}
\textbf{Feature Extraction and Fusion:} We use Facebook's implementation of ResNet152 \cite{he2016deep} to extract multilevel visual features from the input image by taking the output of the two last convolutional layers, $\bm{v_1} \in \mathbb{R}^{2048 \times 14 \times 14}$ and $\bm{v_2} \in \mathbb{R}^{2048}$ where $\bm{v_1}$ represents spatial visual features at each image gird location $G=14{\times}14$ and $\bm{v_2}$ represents the pooled visual features at an image level. We use different combinations of $\bm{v_1}$ and $\bm{v_2}$ in the baseline models that undergo joint embedding with the question features. Semantic feature $\bm{q} \in \mathbb{R}^{2400}$ is generated in a manner similar to \cite{Farazi_2018_BMVC, benyounescadene2017mutan, fukui2016multimodal} where the question is encoded with skip-thought vectors \cite{kiros2015skip} and passed through GRUs. When generating the visual-semantic embedding, we set $t_v{=}t_q{=}310$, $t_e{=}510$ and use two glimpse attention following the literature \cite{Farazi_2018_BMVC, benyounescadene2017mutan}, making the joint embedding $G \times 510$ dimensional. 

\textbf{Exemplar Implementation:} 
% \MRF{Should we send this to supplementary material to avoid further confusion?}\SK{Either we put all implementation details to supp or keep them all, otherwise it will cause suspicion.} 
We store the joint embeddings $\bm{e}$  of randomly selected 10\%  IQ pairs from the training set in $\bm{\xi}$. Our experiments show that such a sub-sampling does not degrade the performance while significantly improving computational efficiency.
%for subsequent processing due to computational limitations. 
To generate the compact embedding or \emph{Soft-key} set $\bm{\kappa}$, Max Pooling is applied on each entry of $\bm{\xi_i} \in \mathbb{R}^{196 \times 510}$ to generate the compressed embedding $\kappa_i \in \mathbb{R}^{\rho}$ for $\bm{\xi_i}$. For our experiments we set $\rho=140$ which was found optimal through our experiments. We represent $\bm{\kappa}$ using a K-D tree data structure. During testing and second stage of training, we query on $\bm{\kappa}$ to find the index of the closest match to the max-pooled $\bm{e}$ by performing k-nearest neighbour search ($k=1$), and get the joint embedding from $\bm{\xi}$ for that index and set as ${\bm{e}}_{\xi}$. 

\textbf{Answer Classifier:} We create the answer set $\mathcal{D}$ with most frequent $2000$ answers from the training set and formulate the VQA task as a multi-class classification problem on the answer set $\mathcal{D} \in \mathbb{R}^{2000}$ following VQA benchmark \cite{antol2015vqa}. The final attended visual-semantic feature representation $\bm{\Tilde{v}}$ is passed through a fully connected layer to project to answer embedding space where softmax cross entropy loss is applied to predict the most probable answer from $\mathcal{D}$.

%-------------------------------------------------------------------------
\subsection{Baseline Models}
\label{sec:baseline_models}
% \MRF{Removed the textbf notation of a,b,c. Its confusion with math symbols}
We first propose three strong baselines that build on the state-of-the-art Tucker fusion technique \cite{Farazi_2018_BMVC,benyounescadene2017mutan} to generate visual-semantic embeddings for VQA. These baselines are: \textbf{(a)} In the \emph{Concatenation Model}, we concatenate $\bm{q}$ and $\bm{v_2}$ and generate the joint embedding $\bm{e}$ by applying multimodal fusion $\Phi$ on $\bm{q}$ and $(\bm{v_2} \oplus \bm{q})$. The joint embedding is used to refine grid level feature $\bm{v_1}$ by applying attention $\bm{\alpha}$. \textbf{(b)} For \emph{Dual Attention Model}, the $\bm{e}$ is generated as $\bm{e} = \Phi(\bm{q}, \bm{v_2})$. This joint embedding generated from pooled image feature is used to apply attention on the grid level image feature $\bm{v_1}$, and is thus called the dual attention model. \textbf{(c)} Finally, the \emph{Grid Attention Model} only uses the grid level visual features $\bm{v_1}$. The joint embedding is generates as $\bm{e} = \phi(\bm{q},\bm{v_1})$ and it is used to apply attention on $\bm{v_1}$ to refine the visual features based on semantic input. The Grid attention model outperforms other baselines in our experiments (Table~\ref{fig:perfomance_drop_ablation}) which  shows its effectiveness in jointly embedding visual-semantic features. Thus our proposed model uses $\bm{v_1}$ to generate the joint embedding of the exemplars $\mathcal{E}$.

% The three baseline models differ in the way they generate joint embeddings. The use of different combinations for visual and semantic features sheds light on the most effective way to project them to a common space for VQA.

%%%%%%%%%%%%%%%%%%%%%%%%%%%%%%%%%%%%%%%
%-----------------------------------------------------------------------
\subsection{Results}
\label{sec:results}
% We compare the existing VQA models with the baselines and our proposed Joint Embedding Exemplar (JEX) model on the OW-VQA dataset and show that our model achieves state-of-the-art-performance. Further we test our proposed model on semantically motivated VQA splits, namely Novel-VQA \cite{ramakrishnan2017empirical} and VQA-CP v1 and v2 \cite{Agrawal_2018_CVPR} and show that it outperforms their proposed models that reason about new semantic concepts (unknown) from an external knowledge base and by visually grounding the question types, respectively.

\textbf{Benchmarking VQA models on OW-VQA:} We benchmark existing VQA models on OW-VQA dataset and report their performance on both versions of our proposed VQA dataset split. From Table \ref{tab:Test_on_zstest}, we can see VQA models that incorporate multimodal (visual-semantic) embedding (\ie pooling \cite{fukui2016multimodal} or fusion \cite{benyounescadene2017mutan}) compared to the models which only use semantic embedding to generate visual attention achieves higher performance in both versions of OW-VQA. Our exemplar based approach further refines the visual attention by transferring knowledge from exemplar set and we report $1.4\%$ and $0.9\%$ overall accuracy gain over the closet state-of-the-art method on both v1 and v2 respectively. Such an improvement without using any external knowledge base (\ie complementary training on Visual Genome \cite{krishna2016visual}, external image and text corpora) and/or model ensemble  justifies our approach of transferring knowledge from exemplars. Furthermore, the accuracy scores of VQA models reported in Table \ref{tab:Test_on_zstest}, drop significantly when evaluated OW-VQAv2 compared to v1 as the IQA triplets in v2 have less semantic bias. It can also be seen that the joint embedding attention models are more robust against semantic bias than semantic attention models (overall accuracy drop of ${\sim}3.5\%$ compared to ${\sim}5.1\%$). This further strengthens our motivation to make use of such joint embedding space which capture highly discriminative multi-modal features.

%%%%%%%%%%%%%%%%%%%%%%%%%%%%%%%%
\begin{table*}[!htp]
 \begin{minipage}[b]{0.3\textwidth}
\centering
  \scalebox{0.8}{
  \begin{tabular}{lcccc}
    \toprule
    	\rowcolor{color3}								& \multicolumn{2}{c}{CPv1}   		   & \multicolumn{2}{c}{CPv2} 	    \\
		\rowcolor{color3}								& All 	& Other & All 	& Other \\ 
	    \midrule
    JEX (Ours)                          & 38.3 & \textbf{43.7}
                                        & \textbf{36.8} & \textbf{41.4}\\								
    \midrule
    GVQA \cite{Agrawal_2018_CVPR}       & \textbf{39.2} & 24.9 & 31.3 & 22.1 \\
    MCB \cite{fukui2016multimodal}      & 34.4  & 39.9  & 36.3  & 40.6	\\
    SAN \cite{yang2016stacked}          & 26.9  & 24.7  & 25.0  & 27.7	\\
    NMN \cite{andreas2016neural}        & 29.6  & 27.9  & 27.5  & 25.7	\\
    VQA \cite{antol2015vqa}             & 23.5  & 17.4	& 19.8	& 14.4	\\
    \bottomrule
  \end{tabular}}\vspace{0.5em}
  \caption{\small Evaluation on VQA-CP dataset \cite{Agrawal_2018_CVPR} along with comparisons.}
  \label{tab:Test_on_CP}
  \end{minipage}
  \hfill
  \begin{minipage}[b]{0.3\textwidth}
    \centering
  \scalebox{0.75}{
  \begin{tabular}{p{2.0cm} cccc}
    \toprule
   					\rowcolor{color3}				& \multicolumn{4}{c}{Novel-VQA} \\
    				\rowcolor{color3}					& All 	& Y/N   & Num.  & Other \\ 
    \midrule
    JEX (Ours)                          & \textbf{55.1} & \textbf{78.7} & \textbf{36.5}
                                        &\textbf{42.4}\\
    \midrule
    Arch-1\cite{ramakrishnan2017empirical}
                                        & 41.8 & 76.6 & 28.5 & 25.7 \\
    Arch-2\cite{ramakrishnan2017empirical}
                                        & 39.9 & 75.9 & 28.9 & 22.8 \\
    VQA \cite{antol2015vqa} & \multirow{1}{*}{39.4} & \multirow{1}{*}{74.0} & \multirow{ 1}{*}{27.5} & \multirow{1}{*}{23.1} \\
    % VQA \cite{antol2015vqa} & 39.4 & 74.0 & 27.5 & 23.1 \\
    VIS+LSTM \cite{ren2015exploring}    & \multirow{ 2}{*}{35.0} & \multirow{ 2}{*}{71.1} & \multirow{ 2}{*}{28.2} & \multirow{ 2}{*}{17.0}\\
    \bottomrule
  \end{tabular}}\vspace{0.5em}
  \caption{\small Evaluation on Novel-VQA \cite{ramakrishnan2017empirical} dataset along with comparisons. %d-LSTM Q + norm I and VQA are the same \cite{antol2015vqa}\SK{We can just rename d_LSTM Q+norm I to VQA}
  }
  \label{tab:Test_on_novel}
  \end{minipage}
  \hfill
  \begin{minipage}[b]{0.34\textwidth}
    \centering
  \scalebox{0.72}{
  \begin{tabular}{lcccc}
    \toprule
    		%	\rowcolor{color3}						    & \multicolumn{4}{c}{VQAv2 Validation-set} \\
				\rowcolor{color3}					VQAv2 Val-set $\rightarrow$	   & All 	& Y/N & Num	& Other \\
	\midrule
    JEX (Ours)                              & \textbf{61.1} & \textbf{79.9}
                                            & \textbf{39.1} & \textbf{52.6}\\
    \midrule
    MUTAN \cite{benyounescadene2017mutan}    & 60.0 & 79.2 & 37.8 & 51.3 \\
    Support-Set \cite{Teney_2018_ECCV}       & 59.9 & - & - & -          \\
    MCB \cite{fukui2016multimodal}           & 59.1 & 77.3 & 36.7 & 51.2 \\
    HieCoAtt \cite{lu2016hierarchical}       & 54.6 & 71.8 & 36.5 & 46.3 \\
    DCN+LQIA\cite{patro2018differential}\footnotemark
                                            & 53.3 & 70.6 & 34.6 & 44.4 \\
    SAN \cite{yang2016stacked}              & 52.0 & 68.9 & 34.6 & 43.8 \\                          
    GVQA\cite{Agrawal_2018_CVPR}            & 48.2 & 72.0 & 31.2 & 34.7 \\
    \bottomrule
  \end{tabular}}\vspace{0.5em}
  \caption{\small Comparison on std. VQAv2\cite{goyal2016making} when model learned on train and tested on val set. }
  \label{tab:Test_on_VQA}
  \end{minipage}
\end{table*}
%%%%%%%%%%%%%%%%%%%%%%%%%%%%%%%

\textbf{Performance drop when evaluated on \emph{Unknown}:} We perform an ablation study to quantify the role of different components of our proposed model on OW-VQA-v2 dataset and compare the performance of our baseline models and full model, along with exemplar-attention-only variant. In this experiment, we train the models on \emph{Trainset} and evaluate on \emph{Valset}, \emph{Valset-known} and \emph{Valset-Unknown} which enables us to perform a comparative analysis on the models' ability to reason about Known and Unknown concepts (see Table~\ref{fig:perfomance_drop_ablation}).
%Fig.~\ref{fig:perfomance_drop_ablation} shows the comparisons of our full model w.r.t to the ablated versions and reports the overall accuracy on Valset, Valset-Known and Valset-Unknown
We also report the number of trainable parameters required for each model. From the bar plot, it can be observed that all model variants achieved higher accuracy on \emph{Valset} compared to \emph{Valset-Unknown} and lower accuracy when compared with \emph{Valset-Known} when trained with only Known concepts. Among the baseline methods, the grid attention variant achieves the highest accuracy with the least number of trainable parameters. Interestingly, when only the joint feature encoding from exemplar (exemplar-attention-only variant) is used, it achieves a relatively reasonable overall accuracy of ${\sim}51\%$. This shows that the exemplar feature indeed encapsulates valuable information for the VQA task. Our full model incorporates both grid attention and exemplar attention with a small increase in the number of trainable parameters, and achieves higher accuracy than the other variants. Furthermore, there is a significant drop in overall accuracy when tested on Unknown concepts compared to Known concepts which signifies the knowledge gap a VQA model needs to address when reasoning about Unknown concepts. In Table. \ref{fig:perfomance_drop_ablation}, note that accuracy difference between Known and Unknown concepts is $6.1$ for JEX which is $12.68\%$ lower than that of the gird attention model. This quantifies the value added by using exemplars to bridge that gap in comprehending Unknown concepts.  

%%%%%%%%%%%%%%%%%%%%%%%%%%%%%%%%%%%%%%%
\footnotetext{Compared with k=1, where only one nearest neighbour was used.}
%%%%%%%%%%%%%%%%%%%%%%%%%%%%%%%%%%%%%%%

\textbf{Evaluation on semantically separated VQA splits:} We evaluate our exemplar based approach on semantically motivated VQA-CP~\cite{Agrawal_2018_CVPR} and Novel VQA~\cite{ramakrishnan2017empirical} datasets where the former separated the challenging semantic concepts in the \emph{testset} and the latter placed least frequent nouns and associated IQA triplets in the \emph{testset}. Although, our motivation is orthogonal and our definition of \emph{Novel Concepts} is heterogeneous to these semantically motivated approaches, we showcase the effectiveness of our exemplar based approach on their settings. In Table~\ref{tab:Test_on_novel}, we compare the performance of JEX model on Novel-VQA split with performances of baseline and proposed methods (Arch-1 and Arch-2) reported in \cite{ramakrishnan2017empirical}. Our exemplar based approach outperforms the best variant of Arch-1 and Arch-2 by $13.3\%$ and $15.2\%$. This is to be noted that even if the proposed approaches by Ramakrishnan \etal \cite{ramakrishnan2017empirical} incorporate external knowledge, both semantic (\ie books) and visual (\ie examples from ImageNet \cite{deng2009imagenet}), our model achieves superior performance by only leveraging information from training examples. 

We also evaluate our model on both versions of VQA-CP dataset and report performance against other benchmarks and their proposed GVQA \cite{Agrawal_2018_CVPR} dataset in Table~\ref{tab:Test_on_CP}. It shows that in VQA-CPv1, GVQA achieves a slightly higher ($0.9\%$)  Overall accuracy than JEX, but performs significantly low ($18.8\%$) compared to JEX for \emph{Other} type questions. %\NB{This last sentence is broken} 
GVQA employs separate question classifiers for Y/N and non-Y/N (\ie Num, Other) questions that account for its high accuracy in Y/N questions which results in higher Overall accuracy. However, when evaluated on VQA-CPv2, JEX outperforms GVQA in both Overall and Other question accuracy by $5.5\%$ and $19.3\%$ respectively because  VQA-CPv2 has a more balanced distribution of question categories and a considerably lower language bias \cite{goyal2016making}. 
% The questions in the Other category require a more complex reasoning in the joint embedding space which our model is able perform more effectively twith $\sim 20\%$ performance gain without explicitly training for separate question categories.

%%%%%%%%%%%%%%%%%%%%%%%%%%%%%%%%%%%%%%%

\textbf{Evaluation on standard VQA setting:}
We evaluate our model on VQAv2 validation set \cite{goyal2016making} and compare its performance with other attention based models. It is worth noting that we only compare with their single model without data augmentation which is similar to our setting for fair comparison.
%due to the difficulty in reproducing the ensemble performance.
%(\ie complementary training on Visual Genome~\cite{krishna2016visual} or other knowledge base)  performance as reproducing the ensemble performance reported is often not possible. 
%\NB{Second half of last sentence is broken} 
From Table \ref{tab:Test_on_VQA} it can be seen that our model outperforms the Tucker decomposition based model by Ben-younes \etal \cite{benyounescadene2017mutan} which has a similar architecture to our baseline models. Further, it also outperforms the Support-Set model proposed by Teney \etal \cite{Teney_2018_ECCV} in a similar setting where the support set contains example representation of question, answers and image. Interestingly, the overall accuracy of GVQA~\cite{Agrawal_2018_CVPR} without an ensemble and/or oracles is $18.9\%$ lower than JEX in a standard VQA setting.

%\MRF{Do we need a summary? Or we can talk about it in the conclusion?}\SK{No we don't, its good you removed that.}
% In summary, our we benchmark our proposed model on OW-VQA dataset where we evaluate a model on previously unknown visual and semantic concept and report state-of-the-art performance. Further, we also exhibit our models superiority on semantically motivated splits of the VQA and the standard VQA setting. This supports our hypothesis that learning to reason from known concepts with exemplars adds value to VQA performance in both general and novel concepts setting.

% higher accuracy on compared to the Valset and lower accuracy  to $60.14$ and $58.9$ overall accuracy on OWv1 and OWv2 Valset (Known+Unknown concepts). On both version the performance increased to a small degree when evaluated only on Valset-Known compared to Valset. However, the performance significantly decreased when evaluated on Valset-Unknown. For OWv1 and OWv2 the performance drop when evaluated on Unknown concepts compared to Known concepts was $6.59$ and $6.92$ respectively.

%=========================================================================
\section{Conclusion}
Existing VQA systems lack the ability to generalize their knowledge from training to answer questions about novel concepts encountered during inference. In this paper, we propose an exemplar-based transfer learning approach that utilizes the closest Known examples to answer questions about Unknown concepts. A joint embedding space is central to our approach, that effectively encodes the complex relationships between semantic, visual and output domains. Given the IQ pair and exemplar embedding in this space, the proposed approach hierarchically attends to visual details and focuses attention on the regions that are most useful to predict the correct answer. We propose a new Open-World VQA train/test split to fairly compare the performance of VQA systems on Known and Unknown concepts. Our exemplar based approach achieves significant improvements over the state-of-the-art techniques on the proposed OW-VQA setting as well as standard VQA setting, which reinforces the notion of transferring knowledge from rich joint embedding space to reason about Unknown concepts. 

{\small
\bibliographystyle{ieee}
\bibliography{egbib}
}
\newpage
\clearpage

%%%%%%%%% TITLE
% \title{\Large \bf \vspace{-1.2em}Supplementary material for `From Known to the Unknown: Transferring Knowledge to Answer Questions about Novel Visual and Semantic Concepts'}

% \maketitle
%\thispagestyle{empty}

% %%%%%%%%% ABSTRACT
% \begin{abstract}
% The supplementary material is arranged as follows. Sec.~\ref{suppsec:dataset_gen} discusses the detailed protocol undertaken to generate the OW-VQA split complementary to Sec.~4 of the main paper. Sec.~\ref{suppsec:dataset_eval} discusses the recommended evaluation protocol for OW-VQA dataset. Sec.~\ref{suppsec:additional_results} contains additional experimental results that are complementary to those reported in Sec.~5 of the main paper.  
% \end{abstract}

%%%%%%%%% BODY TEXT
%=========================================================================
\def\thesection{\Alph{section}}
\setcounter{section}{0}
\section*{\LARGE Supplementary Material}
\vspace{.5cm}

\section{Dataset generation protocol for OW-VQA}
\label{suppsec:dataset_gen}
\begin{figure*}[!htp]
\begin{center}
  \includegraphics[width=.95\linewidth]{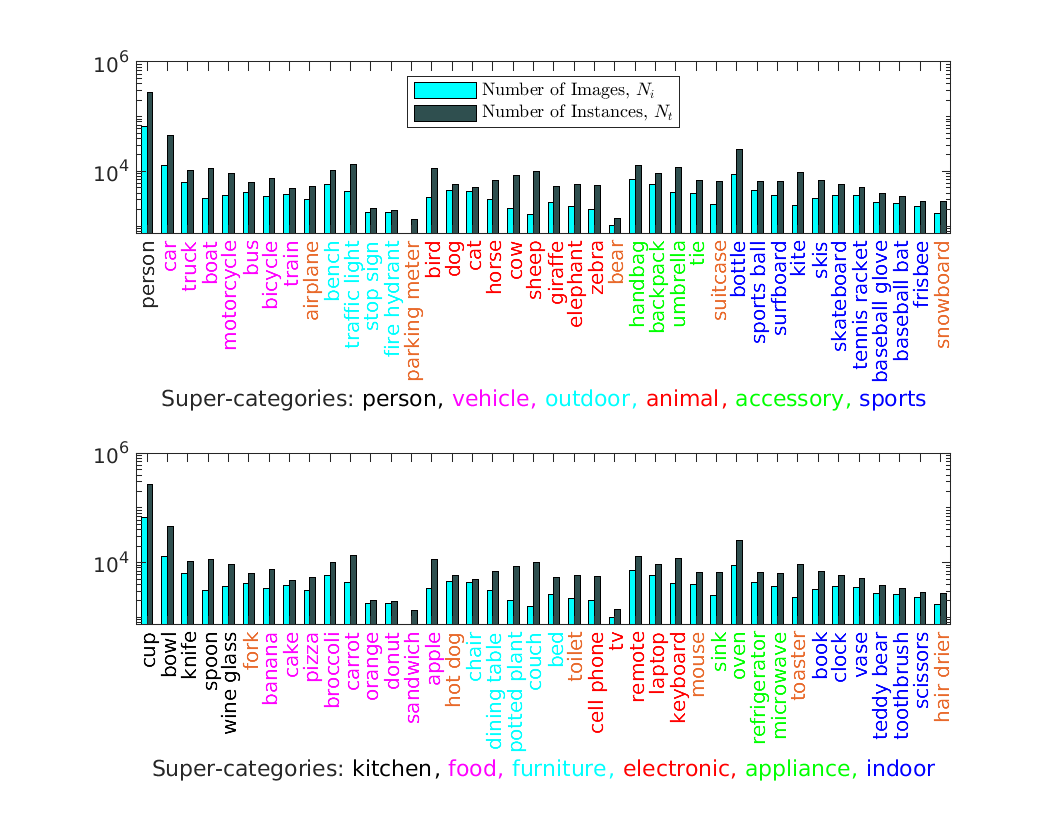}
\end{center}
  \caption{This bar chart shows number of images and number of times each category appears in the MSCOCO Train2014 and Val2014 dataset combined. The super-categories are color-coded and from each super-category one category is selected as \textit{\textcolor{unseenColor}{Unknown}}. 
%   \SK{Change Numner to Number in the legend above.}
  }
\label{fig:image_instance}
\end{figure*}

For each MSCOCO~\cite{lin2014microsoft} category $c$, $N_i$ represents the number of images in which $c$ appears and $N_t$ represents the number of times $c$ appears in the dataset (\ie total instances). These statistics are calculated after merging the MSCOCO Train2014 and Val2014 splits. Fig.~\ref{fig:image_instance} shows $N_i$ and $N_t$ for categories within each super-category of MSCOCO~\cite{lin2014microsoft}. The category names are color-coded to represent the super-category labels and respective {\color{unseenColor}\emph{Unknown}} categories. From this figure, we can see that the categories which appear in the least number of images, the least number of times are selected as Unknown. 

Table~\ref{supptab:dataset_stat} presents statistics of VQA dataset following the proposed Known/Unknown concept separation protocol described in `Image-Question-Answer Split' of Sec.~\ref{sec:dataset_gen} of the main paper. We can see from the statistics that Unknown categories are present in ${\sim}16\%$ of training and validation images. Furthermore, it can be observed, when IQA triplets from the training and validation splits of the VQA datasets are separated on the basis of Known and Unknown concepts, the Unknown IQA triplets also amount to ${\sim}16\%$ of the total. This is an indication that our dataset preparation protocol is able to uniformly separate Known and Unknown concepts even from crowd-sourced, complex, multi-modal dataset like VQA. Such uniform split allows for effective evaluation of a VQA models' ability to reason with Unknown concepts. 

The \emph{Trainset} and \emph{Testset} of the OW-VQA dataset consists of Known and Unknown IQA triplets from corresponding Train splits of VQA datasets respectively. We also propose two validation splits called \emph{Valset-Known} and \emph{Valset-Unknown} from the Val splits of VQA datasets. The \emph{Valset-Known} contains Known IQA triplets and the \emph{Valset-Unknown} contains Unknown IQA triplets from the Valset of respective version. The subdivision of Valset into Known and Unknown splits allows evaluation on both concept types.

\begin{figure}[!b]
\begin{center}
  \includegraphics[width=.9\linewidth]{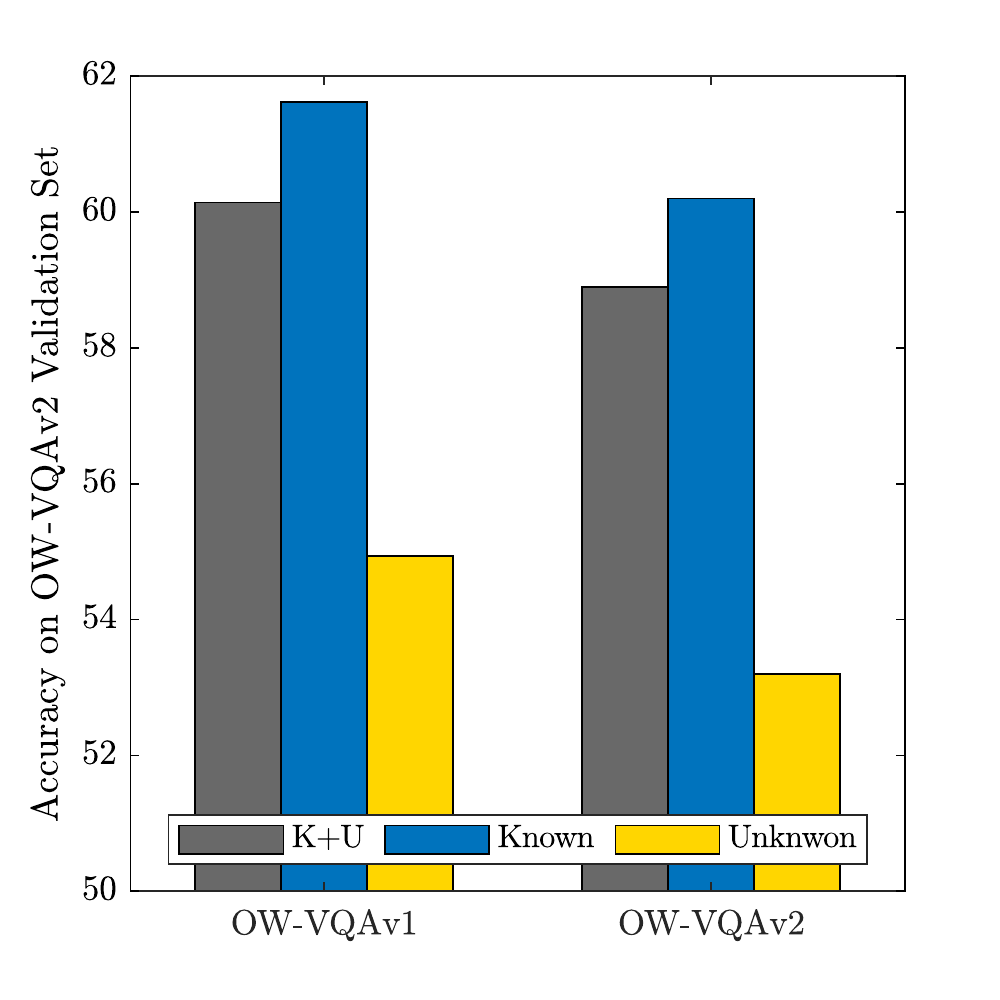}
\end{center}
  \caption{Performance drop when Grid Attention baseline model is trained on Trainset and evaluated on Valset-Known, Valset-Unknown and Valset (K+U) on both versions of OW-VQA dataset.}
\label{suppfig:performance_drop}
\end{figure}

%%%%%%%%%%%%%%%%%%%%%%%%%%%%%%%%%%%%%%%
\begin{table*}[!htp]
  \centering
  \begin{tabular}{l|l|cccc|cccc}
    \toprule
    \multicolumn{2}{c|}{}       & \multicolumn{4}{c}{Training Split}  & \multicolumn{4}{|c}{Validation Split}\\
    \multicolumn{2}{c|}{}
    & Total     & Known      & Unknown    & Unknown\%  & Total     & Known      & Unknown    & Unknown\%\\
    \midrule
    Image   & MSCOCO\cite{lin2014microsoft}
    & 82,783    & 69,557    & 13,226    & 15.98    & 40,504    & 34,137    & 6,367     & 15.72    \\
    \midrule
    \multirow{2}{*}{IQA Tiplet} & VQAv1\cite{antol2015vqa}
    & 224,040   & 187,986   & 36,054    & 16.09    & 120,916   & 101,815   & 19,101    & 15.80     \\
                                & VQAv2\cite{goyal2016making}
    & 402,691   & 336,124   & 66,568    & 16.53    & 213,266   & 178,321   & 34,945    & 16.39     \\
    \bottomrule
  \end{tabular}\vspace{0.5em}
  \caption{Statistics on VQAv1\cite{antol2015vqa} and VQAv2\cite{goyal2016making} dataset following dataset generation protocol described in Sec.~\ref{sec:dataset_gen}.}
  \label{supptab:dataset_stat}
\end{table*}

%%%%%%%%%%%%%%%%%%%%%%%%%%%%%%%%%%%%%%%
%%%%%%%%%%%%%%%%%%%%%%%%%%%%%%%%%%%%%%%
\begin{table*}[!htp]
  \centering
  \begin{tabular}{l|cccc|cccc}
    \toprule
    									& \multicolumn{4}{c|}{VQA-CP v1}   		& \multicolumn{4}{c}{VQA-CP v2} 	    \\
									
    									& All 	& Y/N   & Num.  & Other & All 	& Y/N   & Num.  & Other \\ 
    									\midrule
    JEX (Ours)                          & 38.3  & 43.7 & \textbf{12.5} & \textbf{43.7}
                                        & \textbf{36.8} & 41.4 & \textbf{12.1}& \textbf{41.4}\\
    \midrule
    GVQA \cite{Agrawal_2018_CVPR}         & \textbf{39.2} & \textbf{64.7} & 11.9 & 24.9 
                                        & 31.3  & \textbf{58.0}  & 11.7  & 22.1 \\
    MCB \cite{fukui2016multimodal}      & 34.4  & 38.0  & 11.8  & 39.9  & 36.3  & 41.0  & 12.0  & 40.6	\\
    SAN \cite{yang2016stacked}          & 26.9  & 35.3  & 11.3  & 24.7  & 25.0  & 38.3  & 11.1  & 27.7	\\
    NMN \cite{andreas2016neural}        & 29.6	& 38.8  & 11.2  & 27.9  & 27.5  & 38.9  & 11.9  & 25.7	\\
    VQA \cite{antol2015vqa}             & 23.5  & 34.5  & 11.4  & 17.4  & 19.8	& 34.3	& 11.4	& 14.4	\\
    % per Q-type prior\cite{antol2015vqa}	& 8.39  & 14.70	& 8.34 	& 2.14 	& 8.76 	& 19.36	& 11.70	& 2.39	\\
    % d-LSTM Q\cite{antol2015vqa}         & 20.16 & 35.72	& 11.07	& 8.34 	& 15.95	& 35.09	& 11.63	& 7.11	\\
    \bottomrule
  \end{tabular}\vspace{0.5em}
  \caption{Detailed evaluation on VQA-CP\cite{Agrawal_2018_CVPR} dataset.}
  \label{supptab:Test_on_CP_full}
\end{table*}
%%%%%%%%%%%%%%%%%%%%%%%%%%%%%%%%%%%%%%%

\section{Evaluation protocol for OW-VQA}
\label{suppsec:dataset_eval}
There are two main ways to evaluate a models performance on the proposed OW-VQA dataset.

\textbf{(a)} For the purpose of debugging and running validation experiments, one can train a VQA model on OW-VQA Trainset and evaluate on Valset-Known or Valset-Unknown or the whole Valset. The OW-VQAv1 Trainset contains ${\sim}187k$ IQA triplets, and Valset-Known and Valset-Unknown contains ${\sim}101k$ and ${\sim}19k$ IQA pairs respectively. The OW-VQAv2 has more IQA triplets, where the Trainset contains ${\sim}336k$ IQA triplets, and Valset-Known and Valset-Unknown contains ${\sim}178k$ and $\sim34k$. 
% In Fig.~\ref{suppfig:performance_drop} show evaluation on these validations splits. 

\textbf{(b)} To do a more comprehensive evaluation, it is recommended to train the model on OW-VQA Trainset and evaluate on Testset or Testset+Valset-Unknown, as they have more Unknown IQA pairs than Valset-Unknown. 
% In the later case, validation must not be performed on Valset-Unknown alone. 
For OW-VQAv1 and v2, the Testset contains ${\sim}36k$ and ${\sim}66k$ IQA respectively. When combined with respective Valset-Unknown it presents an even larger setting to evaluate on Unknown concepts. 

\section{Additional results}
\label{suppsec:additional_results}
Fig.~\ref{suppfig:performance_drop} reports the overall accuracy of Grid Attention baseline model trained on Trainset and evaluated on validation splits of OW-VQAv1 and OW-VQAv2. It can be seen that the Known-Unknown accuracy gap is lower in v1 and higher in v2. This is due to the language bias present in VQAv1 dataset and the model used this bias to correctly answer questions about Unknown concepts.

Table~\ref{supptab:Test_on_CP_full} reports the comparison of proposed JEX model and other contemporary VQA models on both versions of VQA-CPv1 and v2 \cite{Agrawal_2018_CVPR}, including accuracy scores of all question categories. It can be seen that GVQA \cite{Agrawal_2018_CVPR} achieved higher accuracy on the Y/N questions than the proposed JEX model. As mentioned in the `Evaluation on semantically separated VQA splits' part of Section~\ref{sec:results} of the main paper, GVQA employs a separate training module for Y/N questions which helps achieve higher accuracy for Y/N questions. However, for all other question categories the proposed JEX model achieved higher accuracy than GVQA. 
% \NB{Not quite clear of the point you are making in this second sentence.  use something like the proposed JEX model instead of it.} \MRF{Updated}

\end{document}